\documentclass[conference]{IEEEtran}
\IEEEoverridecommandlockouts
\usepackage{cite}
\usepackage{amsmath,amssymb,amsfonts}
\usepackage{graphicx}
\usepackage{textcomp}
\usepackage{xcolor}
\usepackage{bm}
\usepackage{algpseudocode}
\usepackage{algorithm}
\usepackage{balance}

\def\BibTeX{{\rm B\kern-.05em{\sc i\kern-.025em b}\kern-.08em
    T\kern-.1667em\lower.7ex\hbox{E}\kern-.125emX}}
\begin{document}

\newcommand{\E}{\mathbb{E}}
\newcommand{\V}{\mathbb{V}}
\newcommand{\m}{\text{m}}
\newcommand{\K}{\mathbf{K}}
\newcommand{\GP}{\mathcal{GP}}
\newcommand{\N}{\mathcal{N}}
\newcommand{\x}{\mathbf{x}}
\newcommand{\X}{\mathbf{X}}
\newcommand{\f}{\mathbf{f}}
\newcommand{\y}{\mathbf{y}}
\newcommand{\I}{\mathcal{I}}
\newcommand{\1}{\mathbf{1}}
\newcommand{\0}{\mathbf{0}}
\newcommand{\R}{\mathbb{R}}
\newcommand{\e}{\mathbf{e}}
\newcommand{\h}{\mathbf{h}}
\newcommand{\Transp}{\mathsf{T}}

\title{Mapping the magnetic field using a magnetometer array with noisy input Gaussian process regression}

\author{\IEEEauthorblockN{Thomas Edridge and Manon Kok}    \IEEEauthorblockA{Delft Center for Systems and Control, \\
Delft University of Technology, Delft, the Netherlands \\        
Email: \{t.i.edridge, m.kok-1\}@tudelft.nl}}
\maketitle

\begin{abstract}
Ferromagnetic materials in indoor environments give rise to disturbances in the ambient magnetic field. Maps of these magnetic disturbances can be used for indoor localisation. A Gaussian process can be used to learn the spatially varying magnitude of the magnetic field using magnetometer measurements and information about the position of the magnetometer. The position of the magnetometer, however, is frequently only approximately known. This negatively affects the quality of the magnetic field map. In this paper, we investigate how an array of magnetometers can be used to improve the quality of the magnetic field map. The position of the array is approximately known, but the relative locations of the magnetometers on the array are known. We include this information in a novel method to make a map of the ambient magnetic field. We study the properties of our method in simulation and show that our method improves the map quality. We also demonstrate the efficacy of our method with experimental data for the mapping of the magnetic field using an array of 30 magnetometers. 
\end{abstract}


\section{Introduction}


Ferromagnetic materials are prevalent in indoor environments and give rise to disturbances in the ambient magnetic field \cite{le20123, kok2018scalable, wahlstrom2013modeling}. In recent years, the disturbances in magnetic field maps are used to improve localisation in indoor environments \cite{solin2016terrain, li2012feasible, robertson2013simultaneous, vallivaara2010simultaneous, viset2022extended, storms2010magnetic}. We make use of a Gaussian process (GP) \cite{williams2006gaussian} to learn magnetic field maps using magnetometers, which measure the ambient magnetic field. We consider the magnetometer locations as the inputs of the GP and the magnitude of the magnetometer measurements as the outputs of the GP. However, the true position of the magnetometer is frequently only approximately known, which reduces the quality of the map. In this paper, we address this shortcoming by using an array of magnetometers. The position of the array is approximately known, but the relative locations of the magnetometers on the array are assumed to be known. The information that magnetometers are placed on an array is extra information that can improve the quality of the magnetic field map \cite{skog2018inertial}.

\begin{figure}[t!]
    \centering
    \includegraphics[width=0.5\textwidth]{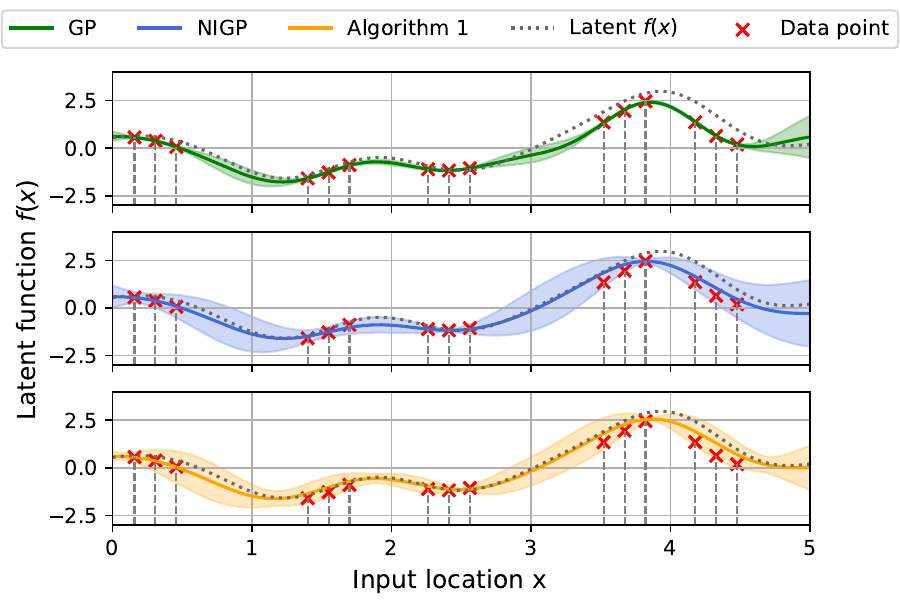}
    \caption{1D example of a synthetic data set where the data is collected using a noisy input array with 3 sensors. The vertical dashed grey lines show the separation of the 3 sensors on the array. We show the regression of a GP (green), NIGP (blue) and our method (orange), all with mean and a confidence interval 
 of 95\%. Data are in red and the latent function $f(x)$ is in grey. The input noise is sampled from a covariance of $\Sigma_\text{x} = 0.05$.}
    \label{fig:IntroductionMeanComparison}
\end{figure}

Typically, GP approaches assume deterministic, known inputs. In the top of Fig. \ref{fig:IntroductionMeanComparison} we see a GP in green which assumes deterministic inputs but uses noisy inputs. This prediction is wrongfully overconfident. We are interested in the case when input locations are only approximately known. Exact inference with noisy inputs is generally intractable 
\cite{mchutchon2015nonlinear, damianou2014variational, girard2002gaussian, candela2003propagation}. One solution uses an approximate analytical approach where the input noise is treated as a function of the input location \cite{goldberg1997regression}. 
However, this method does not take advantage of the shape of the posterior function \cite{bijl2018lqg}. Knowledge of the shape of the posterior function was included in \cite{girard2003learning} for noisy prediction locations, but not for noisy training data. In noisy input GP regression (NIGP) \cite{mchutchon2011gaussian} knowledge of the shape of the posterior function for both training data and prediction locations is included. To be more precise, NIGP maps the input uncertainty to the output via the gradient of the posterior mean. This is illustrated in blue in Fig. \ref{fig:IntroductionMeanComparison}. When the posterior mean function is flat, no input uncertainty gets propagated. When the posterior mean function is steep, the input uncertainty has a large effect. This can be seen in the posterior variance change from GP (green) to NIGP (blue) when $x = 3$. 

We show how we can build magnetic field maps using an array of magnetometers where the array location is only approximately known, but the relative locations of the magnetometers on the array are assumed to be known. To this end, we extend on the work from \cite{mchutchon2011gaussian} and inform our model of the correlation between input locations arising from the fact that the magnetometers are placed on an array. Our method explicitly takes into account that the three sensors in Fig. \ref{fig:IntroductionMeanComparison} are placed on an array. This decreases the posterior variance of our method compared to NIGP, which can be seen in the range from $x = 3$ to $x = 5$. Additionally, the posterior mean of our method more closely resembles the latent function, which can be seen around when $x = 2$ and $x = 4.5$. We show under which conditions the posterior mean function improves and the posterior variance reduces.

\section{Problem formulation}
We use an array of magnetometers to learn a map of the ambient magnetic field. The GP takes the magnetometer locations $\bar{\x}^\text{n}_{i,k}$ as inputs in the navigation frame n. The array moves about in the navigation frame and this frame is fixed with the rotation of the earth. The magnitude of the magnetometer measurements $y_{i,k}$ are taken as outputs. We only have access to the noisy magnetometer locations as inputs
\begin{equation}
    \x^\text{n}_{i,k} = \bar{\x}^\text{n}_{i,k} + \e^\text{n}_{\text{x}}, \quad \e^\text{n}_{\text{x}} \sim \N(\0, \Sigma_\text{x}),
    \label{eq:noisyinput}
\end{equation}
where we have $i = 1,\hdots,m$ magnetometers, $k = 1,\hdots,N$ time steps and $\x_{i,k} \in \R^3$, $\e_{i,k} \in \R^3$. We consider a zero mean GP and kernel function $\kappa(\cdot,\cdot): \R^3 \times \R^3 \rightarrow \R$ 
\begin{equation}
\begin{aligned}
    y_{i,k} &= f(\bar{\x}^\text{n}_{i,k} ) + e_\text{y}, \quad e_\text{y} \sim \N(0, \sigma_\text{y}^2), \\
    f(\cdot) &\sim \GP(\0, \kappa(\cdot, \cdot)),
    \label{eq:GP}
\end{aligned}
\end{equation}
where $f(\cdot)$ is the latent function and $e_\text{y}$ is signal noise. We construct the covariance matrix with the widely used squared exponential (SE) kernel:
\begin{equation}
    \kappa(\x, \x') = \sigma_\text{f}^2 \exp{\left( - \frac{||\x-\x'||^2}{2 l^2} \right)}. \\
\label{eq:SEKernel}
\end{equation}
The hyper-parameters of this kernel are the length-scale $l$ and the signal variance $\sigma_\text{f}^2$. To interpolate or extrapolate a magnetic field map, we can use the GP to make predictions at previously unseen locations $\x_\star$, where we assume $\x_\star$ is deterministic. To this end, we construct kernel matrices $K(X,X) \in \R^{mN \times mN}$, $K(X,X_\star)$ and $K(X_\star, X_\star)$ by evaluating \eqref{eq:SEKernel} for all pairs of data and prediction locations. The predictive posterior is normally distributed with predictive mean $\bm{\mu}_\star$ and covariance $\Sigma_{\star \star}$ as
\begin{equation}
\begin{aligned}
    \f_\star &\sim \N(\bm{\mu}_\star, \Sigma_{\star \star}), \\
    \bm{\mu}_\star &= K^T_\star (K + \sigma^2_\text{y} \I_{mN}  )^{-1} \y, \\
    \Sigma_{\star \star} &= K_{\star \star} - K^T_\star (K + \sigma^2_\text{y} \I_{mN} )^{-1} K_\star.
    \label{eq:GPposterior}
\end{aligned}
\end{equation}
Where we make use of the shorthand notation $f(\x)$ as $f$ and $K(X,X), K(X,X_\star)$ and $K(X_\star,X_\star)$ as $K, K_\star$ and $K_{\star \star}$ respectively. As mentioned in \eqref{eq:noisyinput}, we only have access to the noisy inputs, so we substitute the true inputs with the noisy version $\bar{\x}^\text{n}_{i,k} = \x^\text{n}_{i,k} - \e^\text{n}_{\text{x},k}$. As we can see in \eqref{eq:GPposterior}, the posterior depends non-linearly on true input locations. We cannot solve this in closed form and we will make use of NIGP \cite{mchutchon2011gaussian} for an approximate-analytical solution. 
\section{Magnetometer array GP regression}
    

\subsection{Magnetometer array modelling assumptions}
The locations of magnetometers $i = 1,\hdots,m$ relative to the centre of the array are assumed to be deterministic and are constant in the body frame b. The body frame is fixed to the centre of the array. The magnetometer locations in the navigation frame can be found at every time step $k$ as
\begin{equation}
    \x^\text{n}_{i,k} = \bar{\x}^\text{n}_{c,k} + R^\text{nb}_k \bm{s}^\text{b}_i  + \e^\text{n}_\text{x},
\end{equation}
where $\e^\text{n}_\text{x}$ is defined in \eqref{eq:noisyinput}, $\bar{\x}^\text{n}_{c,k} \in \R^3 $ is the true location of the array's centre, $ \bm{s}^\text{b}_i \in \R^3$ is the translation from the centre of the array to magnetometer $i$ and $R^\text{nb}$ denotes a rotation from the body frame to the navigation frame. We define the matrix $H \in \R^{3m \times 3}$ as 
\begin{equation}
    H =  \1_m \otimes \I_3,
    \label{eq:H}
\end{equation}
where $\otimes$ denotes the Kronecker product, $\1_m$ is a vector of ones of size $m$ and $H$ contains the information on how the input locations are correlated. We take all input locations on the array into the vector for a single time step \mbox{$\bar{X}^\text{n}_k = \begin{bmatrix}
        (\bar{\x}^{\text{n}}_{1,k})^\Transp & \hdots & (\bar{\x}^{\text{n}}_{m,k})^\Transp
    \end{bmatrix}^\Transp \in \R^{3m}$.}
We also define the vector $ S_k^\text{n} \in \R^{3m} $, which contains all the relative magnetometer locations from the array's origin in the navigation frame and is defined as $S_k^\text{n} = \begin{bmatrix} (R^\text{nb}_k \bm{s}^\text{b}_1 )^\Transp \hdots  (R^\text{nb}_k \bm{s}^\text{b}_m )^\Transp \end{bmatrix}^\Transp$. We now have a formulation for the magnetometer locations in the normal frame in terms of the noisy array's centre location:
\begin{equation}
\begin{aligned}
    X^\text{n}_{k} &= \1_N \otimes \bar{\x}^\text{n}_{c,k} + S_k^\text{n}  + \1_N \otimes \e^\text{n}_\text{x}, \\ 
    &= H \bar{\x}^\text{n}_{c,k}  + S_k^\text{n} + H \e^\text{n}_\text{x}, \\
    &= \bar{X}_k^\text{n} + H \e^\text{n}_\text{x}
    \label{eq:ArrayInputLocations}
\end{aligned}
\end{equation}
where $ X^\text{n}_{k}$ and $ \bar{X}^\text{n}_{k}$ are vectors with all the noisy and true magnetometer locations respectively. The previous equation shows that the magnetometer locations for a single time step are correlated as $X^\text{n}_{k} \sim \N(\cdot , H \Sigma_\text{x} H^\Transp)$. 
\subsection{Constructing GP maps using a magnetometer array}
\label{sec:ArrayMaps}
We return to \eqref{eq:noisyinput} and replace the true input locations with the correlated input locations from \eqref{eq:ArrayInputLocations} and our magnetometer measurements to obtain the GP model
\begin{equation}
    \y_k = f(X^\text{n}_k - H \e^\text{n}_\text{x} ) + \e_\text{y}, \quad \e_\text{y} \sim \N(0,  \sigma_\text{y}^2 \I_{m}).
    \label{eq:arraymeasuements}
\end{equation}
where we have taken all the measurements from the array into the vector $\y_k = \begin{bmatrix}
        y_{1,k} & \hdots & y_{m,k}
    \end{bmatrix}^\Transp \in \R^m$ from which we subtract the mean, as we use a GP with a zero-mean function. This results in the output of our GP being the disturbances in the ambient magnetic field.
A closed form solution to the posterior is intractable in \eqref{eq:GPposterior} as the posterior depends non-linearly on the input locations. We find an approximate solution by taking the first-order Taylor expansion around the input locations similar to \cite{mchutchon2011gaussian}. First, we introduce $
\nabla f(\x) = \begin{bmatrix}
    \frac{\partial f(\x)}{\partial x_{1}} & \hdots & \frac{\partial f(\x)}{\partial x_{d}}
\end{bmatrix}^\Transp$ and $\Delta f(X) = \text{blockdiag} \left(
\nabla f(\x_{1}), \hdots, \nabla f(\x_{m}) 
 \right)$ for $m$ input locations. The approximate measurement equation is then found as
\begin{equation}
        \y_{k} \approx f(X^\text{n}_{k}) - \Delta f(X_k^n) ^\Transp H\e^{n}_\text{x} + \e_\text{y}.
\end{equation}
The first-order Taylor approximation allows us to find an expression for the posterior in closed form. By taking the first-order Taylor expansion, we treat the input noise as heteroskedastic output noise by propagating it to the output via the posterior gradient. This also introduces a new variable, the gradient of the posterior mean function which we need to find. We may find the posterior mean gradient for all time steps as
\begin{equation}
    \nabla f(X^\text{n}_{1:N}) = \nabla K^\Transp (K + \sigma^2_y \I_n  
 +  \Sigma(X^\text{n}_{1:N}))^{-1} \y_{1:N}.
    \label{eq:posteriorderivative}
\end{equation}
Where we make use of $X^\text{n}_{1:N} = \begin{bmatrix} (X^\text{n}_{1})^\Transp & \hdots & (X^\text{n}_{N})^\Transp \end{bmatrix}^\Transp$, $\y_{1:N} = \begin{bmatrix}
    \y_1^\Transp & \hdots & \y_N^\Transp
\end{bmatrix}^\Transp$ and the derivative $\nabla K$ of the SE kernel can be found in \cite{mchutchon2015nonlinear}. The posterior gradient can then be used to find the term $\Sigma(X^\text{n}_{1:N})$, which is an additional covariance term introduced by the input noise as
\begin{equation}
    \Sigma(X^\text{n}_{1:N}) = \Delta f(X^\text{n}_{1:N})^\Transp \left( \I_N \otimes \left( H \Sigma_\text{x} H^\Transp \right) \right) \Delta f(X^\text{n}_{1:N}).
    \label{eq:inputcovaraince}
\end{equation}
Estimating the posterior mean derivative is a chicken-and-egg problem. The posterior mean derivative \eqref{eq:posteriorderivative} depends on the additional input uncertainty \eqref{eq:inputcovaraince} which depends on the posterior mean derivative \eqref{eq:posteriorderivative}. We can find a solution by setting the posterior mean derivative to zero, which is effectively the same as doing regular GP regression. We then iterate between finding \eqref{eq:posteriorderivative} and \eqref{eq:inputcovaraince} until a stopping criterion is met. We use the criterion 
\begin{equation}
    \epsilon_j = || \nabla f(X^\text{n}_{1:N})_j - \nabla f(X^\text{n}_{1:N})_{j-1} || < \delta_\epsilon,
    \label{eq:StoppingCriterion}
\end{equation}
where $j$ denotes the iteration number. We then have the predictive posterior mean and covariance as
\begin{equation}
\begin{aligned}
    \bar{\f}_\star &\sim \N(\bar{\bm{\mu}}_\star, \bar{\Sigma}_{\star \star}), \\
    \bar{\bm{\mu}}_\star &= K^T_* \left(  K + \sigma^2_y \I_{mN}   + \Sigma_x(X^\text{n}_{1:N}) \right)^{-1} \y_{1:N}, \\
    \bar{\Sigma}_{\star \star} &= K_{\star \star} - K^T_\star \left(  K + \sigma^2_y \I_{mN}  + \Sigma_x(X^\text{n}_{1:N}) \right)^{-1} K_\star.
    \label{eq:ArrayPosterior}
\end{aligned}
\end{equation}
We summarize our method in Algorithm \ref{alg:ArrayNIGP}.
\begin{algorithm}
\caption{Posterior mean and posterior covariance estimates using a GP with correlated noisy inputs.}\label{alg:ArrayNIGP}
\begin{algorithmic}
\State \textbf{Inputs:} $X^\text{n}_{1:N}$, $\y_{1:N}$
\State \textbf{Outputs:} $\bar{\bm{\mu}}_*,  \bar{\Sigma}_{**}$
\State Set $ \nabla f(X^\text{n}_{1:N}) \leftarrow 0 $
\While{Not converged according to \eqref{eq:StoppingCriterion}} 
    \State Evaluate $  \Sigma(X^\text{n}_{1:N})$ from \eqref{eq:inputcovaraince}.
    \State Evaluate $ \nabla f(X^\text{n}_{1:N})$ from \eqref{eq:posteriorderivative}.
\EndWhile

\end{algorithmic}
\end{algorithm}

\section{Analysis}
We study the properties of our method compared against NIGP and the effects on the quality of the magnetic field map. First, we study the effects of the input uncertainty w.r.t. the length-scale in Section \ref{sec:inputerror}. Subsequently, in Section \ref{sec:arraysize} we study the effect of the length of the array w.r.t. the length-scale. Finally, we study the  case when all measurements are from a single array in Section \ref{sec:SingleArray}.

The effect of the additional input uncertainty term depends on the spatial variation in the data. In the case where there is no spatial variation in the data, the latent function is flat and the derivative is zero. In this case, the additional input uncertainty term has no effect, and Algorithm~1 and NIGP are identical to a regular GP from \eqref{eq:GPposterior}. Another case is when all magnetometers are on a flat part of the function except for one magnetometer. In this case, Algorithm~1 is identical to NIGP.

\subsection{Effect of the input uncertainty on the quality of the map}
\label{sec:inputerror}
We run 100 Monte Carlo (MC) simulations where we sample the data from the SE kernel \eqref{eq:SEKernel} with hyper-parameters $l = 0.5, \sigma_\text{f}^2 = 1, \sigma_\text{y}^2 = 0.001$. We set the input space to $X \in [0, 5]$ and we place 60 array centre locations on a grid over the entire input domain. We use an array with 5 sensors equidistantly placed to get a total of 300 input locations. We sample noise according to \eqref{eq:ArrayInputLocations}, using a variable input covariance ranging from $\Sigma_\text{x} = 0$ to $\Sigma_\text{x} = 0.25$ and add this to the true input locations after which measurement noise is added according to \eqref{eq:arraymeasuements}. 

\begin{figure}[h!]
    \centering
    \includegraphics[width=0.5\textwidth]{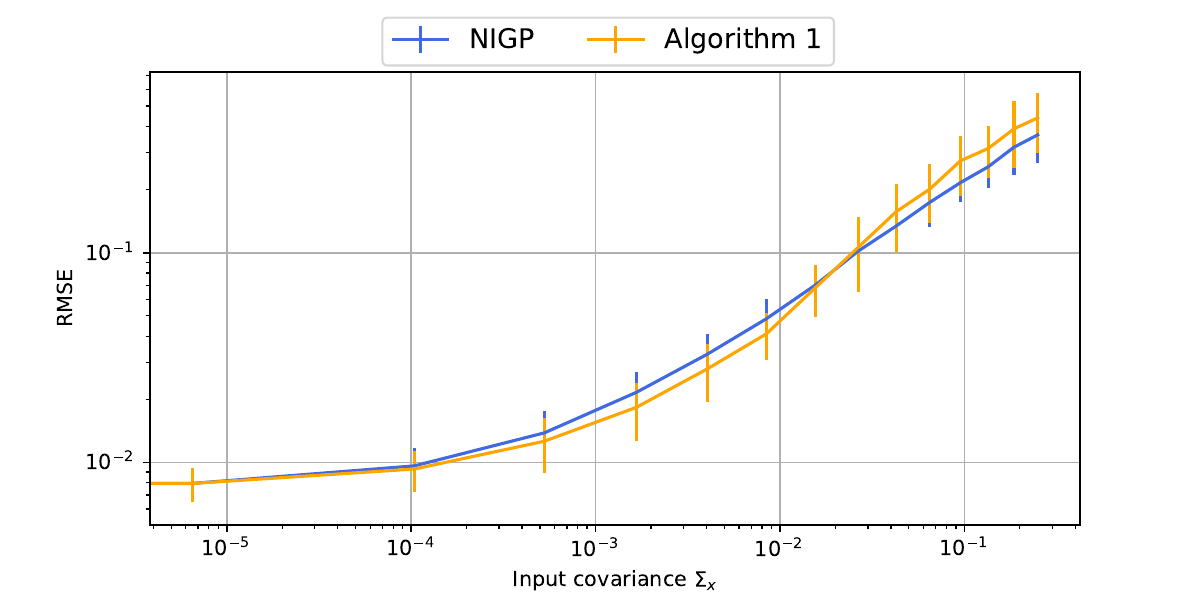}
    \caption{The average and standard deviation of the RMSE over 100 different data sets of Algorithm~1 (orange) and NIGP (blue) as a function of the magnitude of the input uncertainty $\Sigma_\text{x}$.}
    \label{fig:InputUncertainty}
\end{figure}


We compare Algorithm~1 against NIGP and we plot the mean and standard deviation of the RMSE of the 100 different data sets in Fig.~\ref{fig:InputUncertainty}. We show how the average RMSE grows with the input covariance $\Sigma_\text{x}$. As the input covariance increases, from 0 to 0.02, Algorithm~1 has a smaller RMSE than NIGP. We can see that Algorithm~1 provides a better estimate of the posterior mean function compared to NIGP. This makes sense as NIGP omits the correlations between input locations. However, when the input uncertainty grows too large, Algorithm~1 performs worse than NIGP. This can be seen as the input covariance grows larger than 0.02, where the RMSE of Algorithm~1 is compared to NIGP. The reason for this is that when the input error grows too large, we cannot expect the first-order Taylor approximation to be accurate. By introducing correlations between the input locations, Algorithm~1 has fewer degrees of freedom compared to NIGP and thus NIGP is more flexible in modelling the posterior mean resulting in a lower RMSE.

\begin{figure}[h!]
    \centering
    \includegraphics[width=0.5\textwidth]{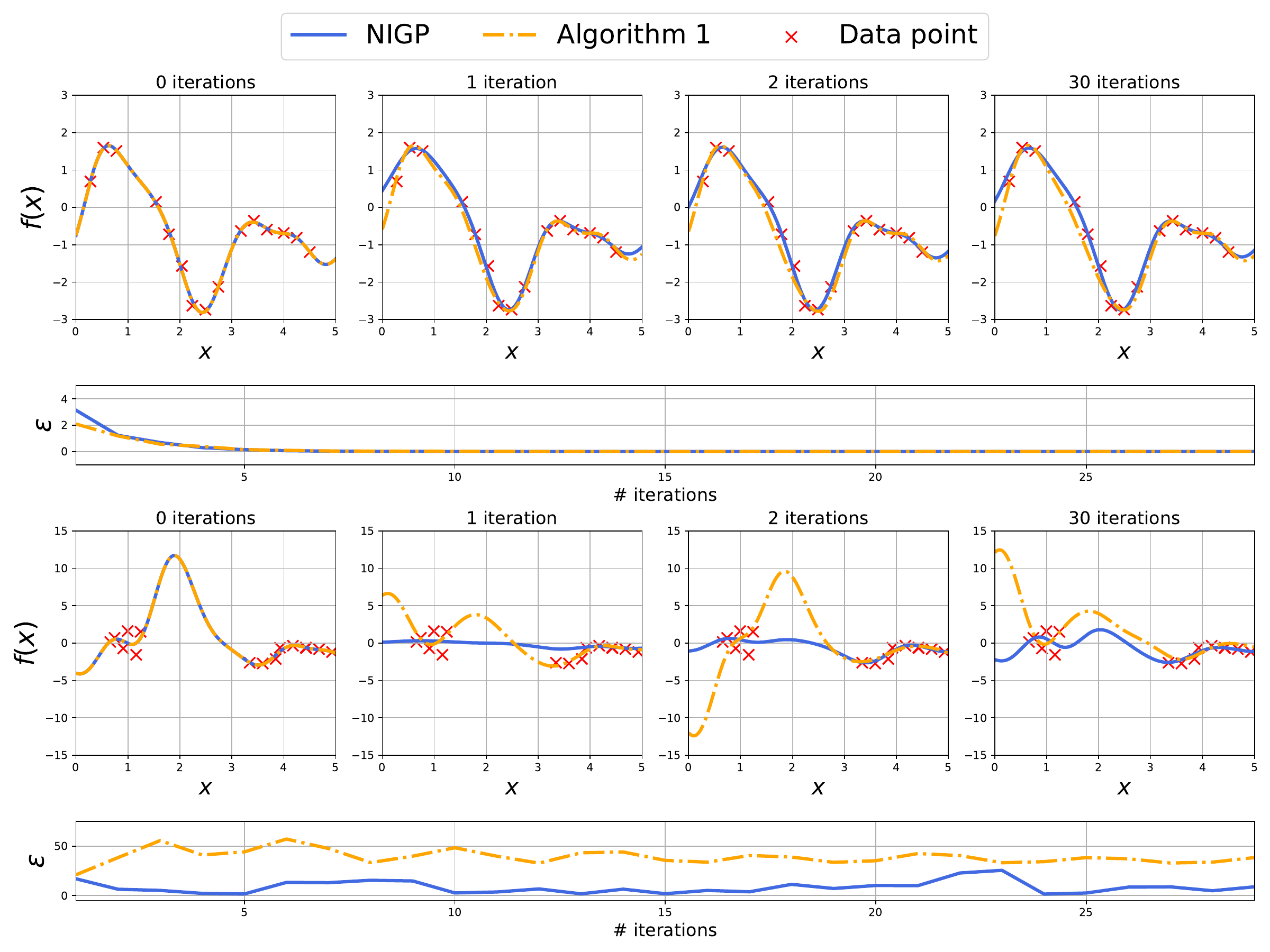}
    \caption{Synthetic data showing 0, 1, 2 and 30 iterations of NIGP and Algorithm~1 along with the error $\epsilon$ from \eqref{eq:StoppingCriterion}. The top row has a small covariance $\Sigma_x = 0.01$ w.r.t. $l = 0.5$ and the bottom row has a large input covariance $\Sigma_x = 0.5$.}
    \label{fig:IterationConvergence}
\end{figure}

Both Algorithm~1 and NIGP iterate to find the additive input covariance $\Sigma(X^\text{n}_{1:N})$ as described in Section \ref{sec:ArrayMaps}. To illustrate the effect of the input error on convergence, we compare Algorithm~1 to NIGP both with a small input covariance of $\Sigma_\text{x} = 0.01$ and a large input covariance of $\Sigma_\text{x} = 0.5$ and a length-scale of $l = 0.5$ for 0, 1, 2 and 30 iterations in Fig. \ref{fig:IterationConvergence}. It can be seen that for $\Sigma_\text{x} = 0.01$, both methods converge within 5 iterations. For $\Sigma_\text{x} = 0.5$, both Algorithm~1 and NIGP do not converge in the 30 iterations. The reason for this is that the input uncertainty is too large w.r.t. length-scale. Data points from different locations start to overlap. This can be seen in the bottom half of the figure around $f(x) = 0, x = 1$. This overlap in data does not adhere to the GP smoothness assumptions, causing Algorithm~1 and NIGP to not converge. NIGP is less affected by the overlap in data points as it assumes no correlations between inputs. It can be observed that NIGP remains relatively flat compared to Algorithm~1, which results in a lower error as data is sampled from a GP with zero-mean assumption. 

\subsection{Effect of the array length on the map quality}
\label{sec:arraysize}
\begin{figure}[h!]
    \centering
    \includegraphics[width=0.50\textwidth]{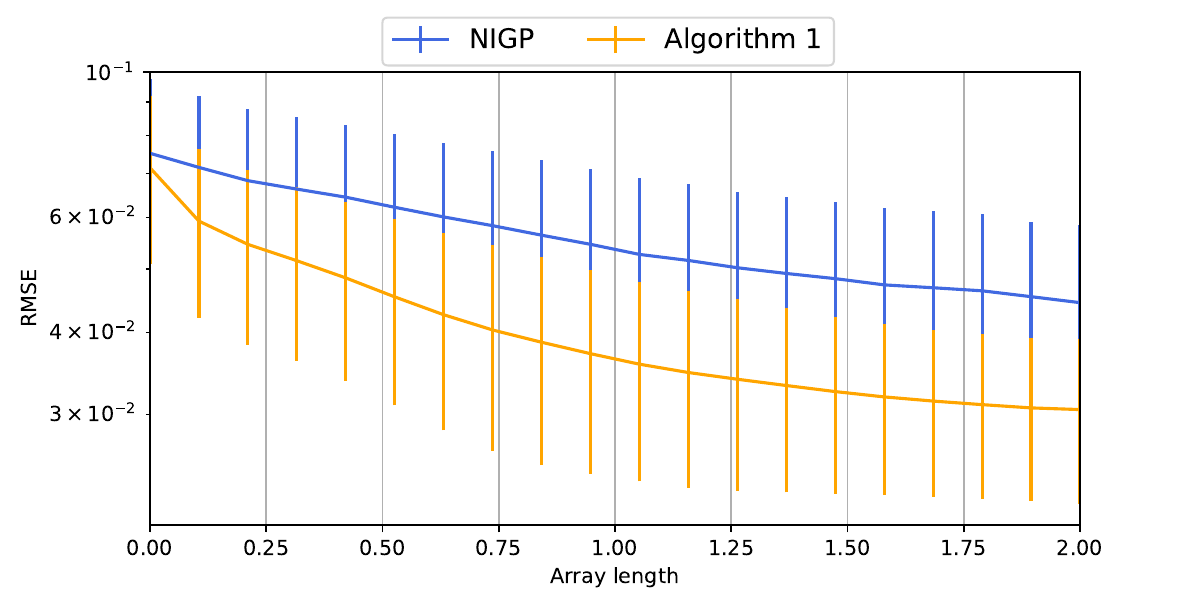}
    \caption{Average RMSE for a simulation of 100 synthetic data sets where we vary the length of an array of 5 sensors w.r.t the length-scale $\l$. The 5 sensors are placed equidistantly along the length of the array. We plot the average predictive posterior RMSE and its standard deviation of NIGP (blue) and Algorithm~1 (orange).}
    \label{fig:ArraySize}
\end{figure}

\begin{figure}[h!]
    \centering
    \includegraphics[width=0.50\textwidth]{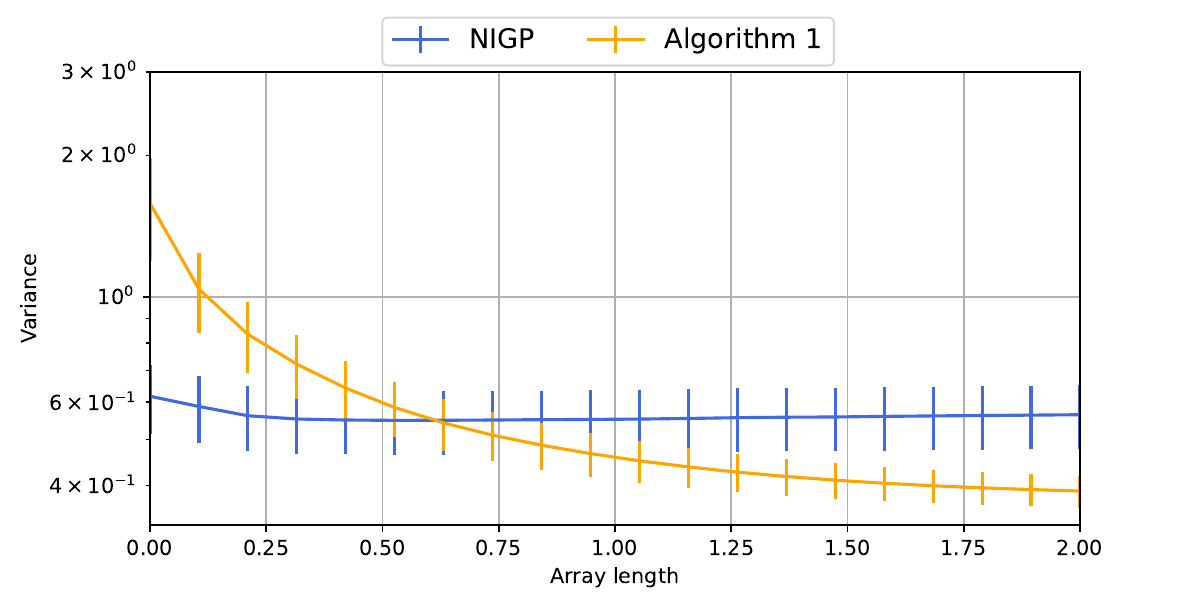}
    \caption{Average posterior variance of a simulation of 100 synthetic data sets where we vary the length of an array of 5 sensors w.r.t the length-scale $\l$. We plot the average and standard deviation of the predictive posterior variance of NIGP (blue) and Algorithm~1 (orange).}
    \label{fig:ArraySizeVariance}
\end{figure}

We also study the effect of the length of the array on the quality of the map. When we increase the length of the array, more spatial variation of the latent function is captured by the array. We run the simulation with the same settings as in Section \ref{sec:inputerror}. We set the input covariance to $\Sigma_\text{x} = 0.05$ and vary the length of the array from 0 to 2. We plot the RMSE and the posterior predictive variance in Figs. \ref{fig:ArraySize} and  \ref{fig:ArraySizeVariance}, respectively. In Fig. \ref{fig:ArraySize} we observe that if the array is of length 0, i.e. the sensors of the array are placed in the same location, Algorithm~1 has a smaller RMSE compared to NIGP. The reason for this is that NIGP omits the correlation terms between input locations, treating each measurement as separate, leading NIGP to be overconfident. For an array length of 0.5 (the size of the length-scale), both Algorithm~1 and NIGP have an RMSE of 0.064 and 0.048 respectively. Algorithm~1 improves more compared to NIGP. The reason for this is that when the length of the array is bigger, Algorithm~1 is able to capture more spatial information. If we increase the array length to 2, the RMSE of Algorithm~1 and NIGP decrease to 0.030 to 0.044 respectively. The decrease in RMSE corresponds to improving the magnetic field map quality. The bigger the size of the array, the better the map quality. 

It can be seen in Fig. \ref{fig:ArraySizeVariance} that when the array length is zero, Algorithm~1 has a larger variance than NIGP. This suggests that NIGP overfits as the variance is smaller but the RMSE is bigger compared to Algorithm~1. The variance of Algorithm~1 and NIGP intersect around an array size of 0.63, just bigger than the length-scale $l = 0.5$. When the array length is 2, the variance of Algorithm~1 decreases to 0.38. If the array size is larger than the length-scale, NIGP starts to underfit as the RMSE is larger and the predictive posterior variance is also larger compared to Algorithm~1. Overall, we see that when the array length increases, the posterior variance deceases. This in line with the results shown in \cite{skog2018magnetic}, where they derive the Cram{\'e}r-Rao lower bound for estimating the position. They show that the positional error decreases as the size of the magnetometer array increase. 

\subsection{Effects when all measurements are from a single array}
\label{sec:SingleArray}
To illustrate the extreme case, we create a data set where all the measurements are from a single array. We omit measurement noise and add an input error of $\e_\text{x} = 0.5$. 

\begin{figure}[h!]
    \centering
    \includegraphics[width=0.50\textwidth]{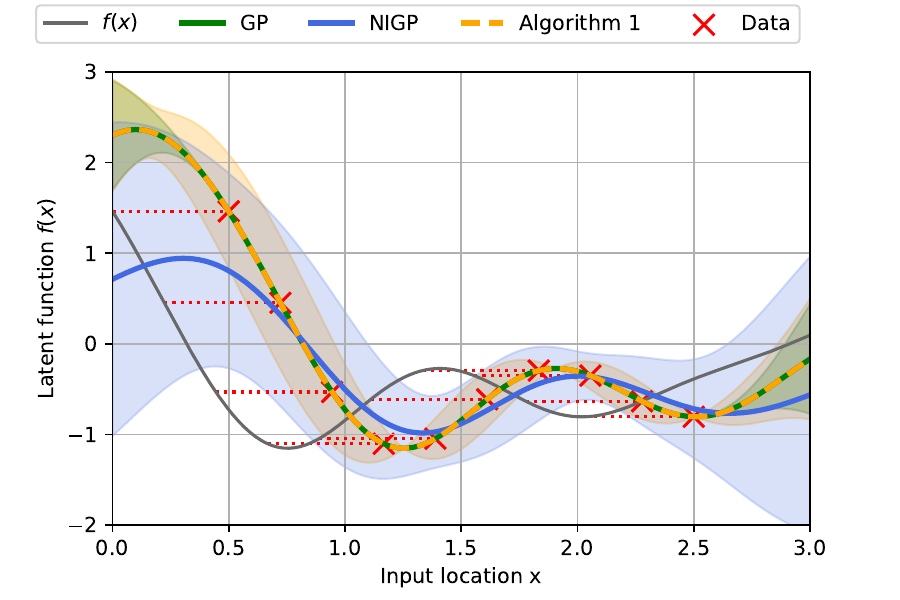}
    \caption{Synthetic data where 10 measurements are taken with a single array. We plot the posterior mean of a regular GP (green), NIGP (blue) and Algorithm~1 (orange) each with 95\% confidence interval.}
    \label{fig:FullArray}
\end{figure}

In Fig. \ref{fig:FullArray} we have 10 data points all captured with a single array. We can see that both Algorithm~1 and a regular GP (which assumes deterministic inputs but uses noisy inputs) are identical in their posterior mean function and they are able to capture the spatial variation of the latent function. In fact, over the span of the array, from  $x = 0.5$, to $x = 2.5$, the posterior mean functions are identical to the latent function shifted in the input direction with the input error $e_\text{x}$.

\section{Experiments}
\begin{figure}[h!]
    \centering
    \includegraphics[width=0.30\textwidth]{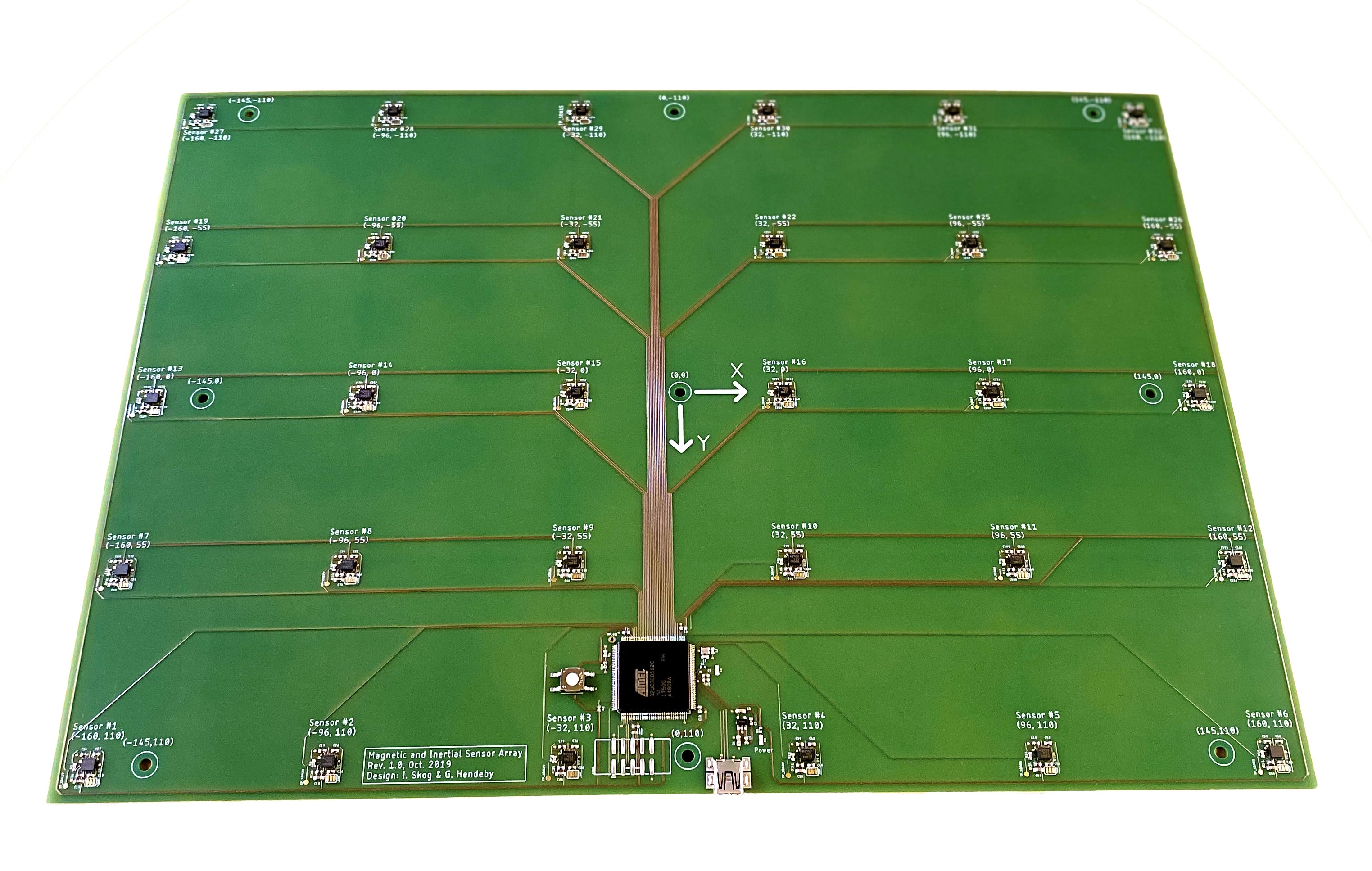}
     \caption{Magnetometer array of dimensions $0.32 \text{m} \times 0.22 \text{m}$ with 30 magnetometers used in the collection of the two data sets. The magnetometers are RM3100 from PNI running at $500 \text{Hz}$.}
    \label{fig:MagnetometerArray}
\end{figure}

Experimental data is collected using an array of size $0.32 \text{m} \times 0.22 \text{m}$ with 30 magnetometers placed on a grid as shown in Fig. \ref{fig:MagnetometerArray}. The ground truth position of the array and its orientation over time are measured using an optical motion capture system (OptiTrack).

\begin{figure}[h!]
    \centering
    \includegraphics[width=0.5\textwidth]{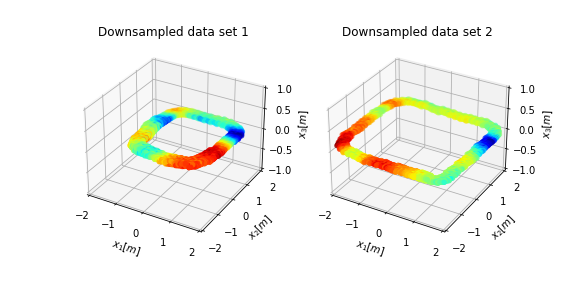}
    \caption{Subsets of data set 1 and 2 showing 1200 measurements of a magnetometer array, where a magnetometer array moves close to the ground in a small square (left) and a big square (right). The colour indicates the value of the measurement.}
    \label{fig:trajectory}
\end{figure}

Two data sets are collected using the array. In the first data set the array moves in a small square (of approximately size $3\text{m} \times 3\text{m}$). In the second data set in a big square (of approximately size $4\text{m} \times 4\text{m}$). Because the computational complexity of GP regression scales cubically with the number of data points, we downsample both data sets to 50 smaller subsets each. Each subset has 1200 data points. In Fig. \ref{fig:trajectory} we show an example of one subset for data set 1 and 2.

\subsection{Magnetometer calibration}
To find the outputs of our GP model, we first calibrate each magnetometer for their bias (offset) $\bm{b}_i$ and distortion matrix $D_i$, where $i = 1, \hdots, m$ according to the initial parameter estimate of \cite{kok2016magnetometer}. This ensures that all magnetometers would provide the same measurements of the ambient magnetic field if they would be placed at exactly the same location. We then consider the magnitude of the calibrated magnetometer measurements $y_{i,k}$ as the output of our GP model
\begin{equation}
    y_{i,k} = || D_i^{-1} \left( \hat{\y}_{i,k} 
 - \mathbf{b}_i \right) ||_2^2,
\end{equation}
where $\hat{\y}_{i,k}$ are the magnetometer measurements. We consider $y_{i,k}$ as the noisy measurements of the underlying magnetic field. We then subtract the mean from all the measurements.

\subsection{Results}
We find the hyper-parameters of the GP for the two data sets by minimising the negative log marginal likelihood according to \cite{williams2006gaussian} using the ground truth position and the measurements of a single magnetometer. The hyper-parameters are found to be $l = 5.59 \times 10 ^{-1}, \sigma_\text{f}^2 = 4.90  \times 10^{-3}, \sigma_\text{y}^2 = 5.31 \times 10 ^{-4}$ and $l = 5.51 \times 10 ^{-1}, \sigma_\text{f}^2 = 4.93  \times 10^{-3}, \sigma_\text{y}^2 = 3.89 \times 10 ^{-4}$ for data sets 1 and 2, respectively.

For each of the 50 subsets of 1200 data points, we individually sample input errors for each array location according to \eqref{eq:ArrayInputLocations}. We sample input errors with a covariance of $\Sigma_\text{x} = \N(\0,  \text{diag}(\beta, \beta, 0))$, where $\beta$ is a scaling factor that ranges from 0.01 to 0.2. We don't add input noise to the vertical axis as we, primarily, move along the two horizontal axes as shown in Fig.~\ref{fig:trajectory}.

\begin{figure}[h!]
    \centering
    \includegraphics[width=0.50\textwidth]{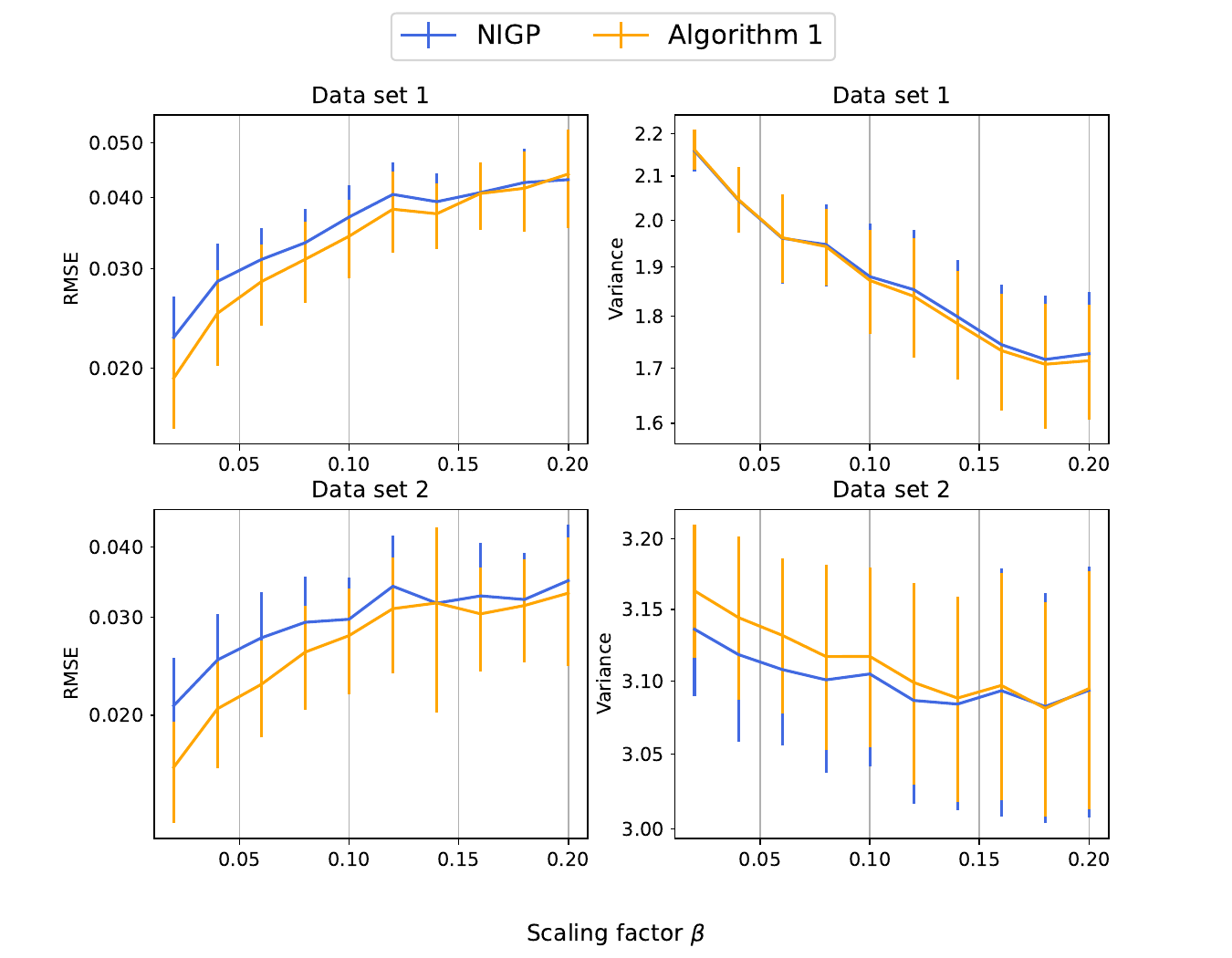}
    \caption{Average RMSE (left) and trace of the posterior variance (right) of a simulation of the the 50 subsets for both data set 1 (top) and data set 2 (bottom). The RMSE is the error from a GP using the true input locations. We compare NIGP (blue) and Algorithm~1 (orange) all of which take noisy inputs. Input noise is sampled for each MC simulation from $\Sigma_\text{x} = \N(\0, \text{diag}(\beta, \beta, 0))$, where $\beta$ is a variable scaling factor.}
    \label{fig:MagnetometerArrayInputUncertaintyRMSE}
\end{figure}

We compare Algorithm~1 and NIGP to a reference GP that uses the true input locations, which acts as a baseline. On the left side in Fig. \ref{fig:MagnetometerArrayInputUncertaintyRMSE} we show the average RMSE of the 50 subset for data set 1 and 2. We can see that when $\beta = 0.01$, the average RMSE of Algorithm~1 is the lowest for both data sets. As $\beta$ grows to 0.2, the RMSE of Algorithm~1 gets closer to that of NIGP. For data set 1 the RMSE of Algorithm~1 and NIGP of equal magnitude at $\beta = 0.16$ and for data set 2 they become similar from $\beta = 0.14$. This is in line with our results from Fig. \ref{fig:InputUncertainty} where we show that Algorithm~1 performs better than NIGP if the error is small enough w.r.t. to the length-scale. 

On the right side in Fig. \ref{fig:MagnetometerArrayInputUncertaintyRMSE} we show the average trace of the posterior predictive variance of the two data sets. Interestingly, we can see that the variance decreases as we increase $ \beta$. As shown in Fig. \ref{fig:trajectory}, we have collected the data along a fixed path. By increasing the input noise, the noisy input locations become more scattered across our prediction domain (i.e. away from the fixed path). This results in the GPs wrongfully lowering the predictive variance in places where in reality no measurement was taken. We also see that the variance of Algorithm~1 has approximately the same magnitude as NIGP. We saw a similar effect in Fig. \ref{fig:ArraySizeVariance}. The array is of size $0.32 \text{m} \times 0.22 \text{m}$ where the length-scale of size $l \approx 0.5$ for both data sets. To decrease the variance compared to NIGP, we need an array of size larger than the length-scale. 

\begin{figure}[h!]
    \centering
    \includegraphics[width=0.550\textwidth]{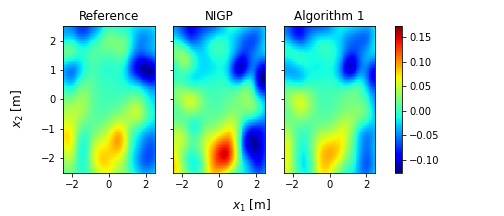}
    \caption{Visual example of the map quality of one of the downsampled data sets from data set 2. The absolute value of the disturbances in the local magnetic field is plotted over an area of $ 5 \times 5 \text{m}^2$. NIGP (middle) and Algorithm~1 (right) are compared to a GP using the true input locations (left) as a reference map. The input error is sampled from $\Sigma_x = \N(\0, \text{diag}(0.05, 0.05, 0))$.}
    \label{fig:MagneticFieldArray}
\end{figure}

In Fig. \ref{fig:MagneticFieldArray} we illustrate the difference in map quality between Algorithm~1 (right) and NIGP (middle) compared to a GP using the true input locations (left). Around locations $\x = \begin{bmatrix} 0 & -2\end{bmatrix}^\Transp$ and $\x = \begin{bmatrix} -2 & 1\end{bmatrix}^\Transp$ Algorithm~1 looks more similar to the reference map compared to NIGP. This shows that a lower RMSE as shown in Fig. \ref{fig:MagnetometerArrayInputUncertaintyRMSE} results in a higher map quality. 


\section{Conclusion and future work}
In this paper, we show that by using an array of magnetometers, we can increase the quality of the magnetic field map. Extending on NIGP \cite{mchutchon2011gaussian}, our method uses the information that the magnetometers are placed on an array. We show that compared to NIGP, our method decreases the RMSE (in comparison to the true latent function) of the predictive posterior mean if the input error is small enough w.r.t. the length-scale. We also show that if the array size is small w.r.t. the length-scale, our method compensates by increasing the posterior variance compared to NIGP. If the array size is big w.r.t. the length-scale, our method reduces the variance compared to NIGP. Finally, we show that our method improves the map quality using experimental data set with an array of 30 magnetometers. An interesting direction of future work is how our method can become more scalable whilst preserving the correlated covariance structure.

\section*{Acknowledgment}
This publication is part of the project “Sensor Fusion For Indoor Localisation Using The Magnetic Field” with project number 18213 of the research program Veni which is (partly) financed by the Dutch Research Council (NWO).

The experimental data used in this paper was collected together with Gustaf Hendeby and Chuan Huang from Link\"oping University. We thank them for allowing us to use this data, and the magnetometer array photo, and for supporting us with the data pre-processing.


\balance
\bibliographystyle{unsrt}
\bibliography{references.bib}

\vspace{12pt}
\color{red}

\end{document}